\documentclass[letterpaper]{article}
\usepackage[preprint]{aaai2027}
\usepackage[hyphens]{url}
\usepackage{graphicx}
\usepackage{natbib}
\usepackage{caption}
\usepackage{amsmath,amssymb,booktabs,array}
\usepackage{xcolor}
\usepackage[most]{tcolorbox}
\usepackage{float}
\usepackage[most]{tcolorbox}
\title{Do ``True'' and ``False'' Add Up to One? Complement Consistency in LLM-Agent Opinion Simulations}

\author{
    Anouar Imel,
    Jin Qi
}

\affiliations{
    School of Information and Communication Engineering\\
    University of Electronic Science and Technology of China (UESTC)\\
    Chengdu, China\\
    202524010105@std.uestc.edu.cn, jqi@uestc.edu.cn
}

\begin{document}
\raggedbottom
\maketitle
\begin{abstract}
In LLM-agent opinion simulations, each agent reports the probability that a claim is true after seeing its own previous probability and the probabilities reported by its neighbors, and that number becomes its state in the next round. If the number describes the agent's position rather than the sentence it was shown, then rewriting the claim as its complement and replacing every probability $p$ with $1-p$ should return the complementary answer, so the two reports should sum to one. We test this with three small open models: Qwen3-4B, Llama-3.2-3B and Gemma-3-4B. On 2,304 agent states taken from a running network simulation, Qwen's paired reports sum to 1.18 on average, and on 48 real fact-checking claims from AVeriTeC they again exceed one. A controlled benchmark of 24 worlds, each with exactly two possible outcomes, shows the same failure in all three models. The discrepancy is two to ten times larger than the change produced by a same-meaning paraphrase, and it persists under repeated sampling and across peer group sizes and report strengths. When the reports are fed back into two populations that start from complementary states, the populations do not remain mirror images: their divergence is positive after five rounds for every model. Averaging each answer with the complement of its paired answer restores consistency by construction and lowers Brier loss averaged over the two wordings, but at the final round it beats the better of the two original wordings for only one of the three models. A schedule-by-targeting experiment on the same model gives a small, stable interaction because such sweeps keep the claim wording fixed. Consistency under rewording therefore needs its own test before simulated probabilities are read as beliefs.
\end{abstract}

\section{Introduction}
A language-model agent can report the probability of a claim after observing its current probability, its neighbors' reports, and source evidence. In a numerical agent simulation, that answer becomes the agent's next state and enters its neighbors' subsequent requests. A report therefore serves two roles: it measures the current response and helps determine the information available in the next round. Conclusions about evidence delivery and collective behavior depend on this feedback process.

Varying graphs, seeds, initial conditions, and update schedules tests how collective outcomes depend on simulation design. A separate question concerns the local update itself: does it assign compatible probabilities to equivalent descriptions of the same information? An intervention can look robust across schedules while the agents assign incompatible probabilities to complementary descriptions, so robustness to design choices leaves the question open.

\begin{figure*}[t]
\centering
\includegraphics[width=\textwidth]{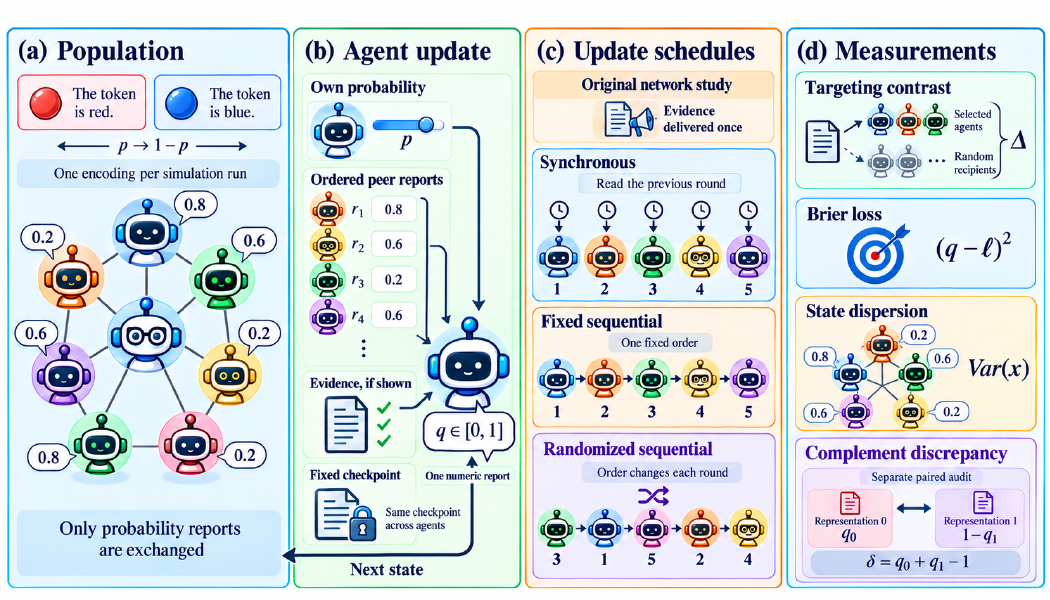}
\caption{Numerical agent experiments. \textbf{(a)} Agents exchange scalar probability reports on a fixed graph; values and network sizes are schematic. \textbf{(b)} Own state, ordered peer reports, and available evidence enter a fixed checkpoint, which returns one number for the next state. \textbf{(c)}The network study compares three schedules with one-time evidence delivery with one-time evidence delivery. Sequential agents can read already-updated neighbors; displayed identities illustrate order. \textbf{(d)} Targeting contrasts, factual loss, and dispersion are separate outcomes. Complement discrepancy comes from separate paired-request studies: the schedule sweep holds the claim wording fixed. The token example illustrates the controlled benchmark, not the campus claims.}
\label{fig:overview}
\end{figure*}

The question can be tested directly (Figure~\ref{fig:overview}). Let $c$ be a binary claim, $p$ the agent's current probability, and $\mathbf r$ its ordered peer reports. Replacing $(c,p,\mathbf r)$ by $(\bar c,1-p,\mathbf 1-\mathbf r)$ expresses the same information relative to the exhaustive alternative $\bar c$, with factual evidence unchanged. Reports that depend on the information rather than on its wording should sum to one. Any discrepancy changes the numerical state supplied to later updates, so a local wording effect can lead to different simulation trajectories.

We connect the local criterion to simulation outcomes through three levels of evidence. A network experiment compares degree-targeted and random evidence delivery under three update schedules. Paired probes then examine agent states encountered in that experiment and negated versions of real fact-checking claims. A controlled benchmark defines exact binary alternatives and transforms the complete request, separating complement discrepancies from paraphrase sensitivity, explicit negation, and variation across repeated samples.

The controlled benchmark also makes feedback directly testable. Two populations start from complementary states, receive the same evidence, and evolve with the same model. Mapping their reports to common coordinates shows whether the paths remain equivalent. Separate repaired populations test whether enforcing consistency improves factual loss once the repair itself changes future requests. Varying the number of peers and the strength of their reports shows how these observations depend on the social information the agent sees.

We make three contributions. First, we define a complement-consistency test
for the update step of an LLM-agent simulation and show that three small open
models fail it on states from a running simulation, on real fact-checking
claims, and on a controlled benchmark in which every world has exactly two
outcomes. Second, we show that the failure exceeds sensitivity to a
same-meaning paraphrase, persists under repeated sampling and across peer
counts and report strengths, and carries into later rounds when reports are
fed back into the simulation. Third, we evaluate averaging each report with
the complement of its paired report, which restores consistency by
construction, and show that its effect on factual loss under feedback depends
on the model.

\section{Related Work}
\paragraph{Probability judgments and logical consistency.}
\citet{betz2023} study probabilistic coherence and evidence integration in language models. \citet{zhu2024} examine probability identities and repeated numerical judgments. Logical consistency also supplies evaluation signals when ground truth is difficult to assess \citep{fluri2023}. These studies motivate checking elicited numbers against relationships implied by the questions.

\paragraph{Wording and response bias.}
Meaning-preserving changes to prompt wording and format can move model
outputs by large margins \citep{sclar2024}, and models respond to
acquiescence-style manipulations of survey items differently from people
\citep{tjuatja2024}. Our test differs in that one transformation changes the
claim, the agent's probability and every peer report together, so the
correct response is fixed by arithmetic.

\paragraph{Consistency repair and elicitation.}
\citet{paleka2025} evaluate logically related forecasts, including negation and paraphrase, and introduce inference-time coherence repair. \citet{andrews2026} examine Dutch-book coherence and elicitation variation. We use the elementary complement-symmetrization identity as an evaluation intervention. Its algebraic guarantees apply to a fixed pair of requests; our empirical question is how the repaired states affect subsequent requests and factual loss under feedback.

The elicited quantity matters. Verbalized confidence and conditional token probabilities are different measurements \citep{tian2023}. Here the output is a generated number about a displayed proposition, not its token-generation probability. Semantic and negation consistency have also been studied for language-model responses \citep{jang2023}; our same-polarity control preserves all numerical inputs, whereas complementation changes every claim-relative probability.

\paragraph{Agent simulation and social influence.}
Networked language agents have been used to study opinion dynamics \citep{chuang2024}, and peer-pressure experiments report direction-dependent opinion change \citep{mehdizadeh2025}. Scheduling is an established modeling choice in agent-based simulation \citep{huberman1993,lawson2000}. Our numerical simulator connects these concerns to wording dependence in the stored state. Averaging numerical reports also differs from self-consistency reasoning over diverse sampled reasoning paths \citep{wang2023}: repeated greedy calls here serve as a repeatability and cost control.

\section{Update Requests and Consistency}
\subsection{The Complement Test}
Let $c$ be a binary proposition about a fixed world, referent, and time. An agent sees its prior probability $p\in[0,1]$, an ordered list of peer probabilities $\mathbf r=(r_1,\ldots,r_k)$, factual evidence $E$ or its absence, and an audience instruction $a$, which may be absent. Write the request and its update as
\begin{equation}
 s=(c,p,\mathbf r,E,a),\qquad q=F_\theta(s;\xi)\in[0,1].
 \label{eq:update}
\end{equation}
Here $\theta$ fixes the model, prompt template, numerical format, and decoding settings; $\xi$ denotes decoding randomness. The world definition is held fixed throughout a pair. Under greedy decoding we write $F_\theta(s)$ for brevity.

Let $\bar c$ be the exhaustive alternative to $c$. The complement transformation is
\begin{equation}
 T(s)=(\bar c,1-p,\mathbf 1-\mathbf r,E,a).
 \label{eq:transform}
\end{equation}
The vector $\mathbf 1-\mathbf r$ complements each peer probability while preserving peer identity and order. Evidence facts, reliability, dates, and unrelated measurements remain unchanged. Evidence supporting $c$ consequently counts against $\bar c$ without being rewritten. The evaluation label $\ell\in\{0,1\}$ becomes $1-\ell$; labels are never shown to the agent. We call the two descriptions the original ($o=0$) and complementary ($o=1$) orientations.

The transformation must be involutive on the defined state: $T(T(s))=s$. This is a condition on the proposition and the numbers, not a demand that English double negations reproduce identical strings. The controlled materials define the two predicates explicitly.

A deterministic updater satisfies complement consistency when
\begin{equation}
 F_\theta(T(s))=1-F_\theta(s).
 \label{eq:consistent}
\end{equation}
For a paired request, define signed and absolute discrepancies as
\begin{align}
q_0(s)&=F_\theta(s), & q_1(s)&=F_\theta(T(s)),\notag\\
\delta(s)&=q_0(s)+q_1(s)-1, & A(s)&=|\delta(s)|.
 \label{eq:discrepancy}
\end{align}
Positive $\delta$ means the two displayed-claim probabilities sum to more than one; negative $\delta$ means they sum to less. We form $A$ within each pair before averaging, so opposing signed errors cannot cancel. Separate model calls create separate contexts, so probability axioms alone do not force their reports to agree. Complement consistency is the operational requirement for reading the reports as states that do not depend on wording; it is distinct from calibration and from full probabilistic coherence.

For each orientation $o$, let $q_o^{\rm base}$ and $q_o^{\rm para}$ be the reports for the base sentence and for a same-meaning paraphrase. Define

\begin{equation}
 H(s)=\tfrac12\bigl(|q_0^{\rm base}-q_0^{\rm para}|+|q_1^{\rm base}-q_1^{\rm para}|\bigr).
 \label{eq:paraphrase}
\end{equation}
The matched comparison averages the complement discrepancies of the two wordings and compares that value with $H$ at otherwise identical factors. It is a descriptive comparison and does not assume that the two linguistic transformations are equally difficult.

\subsection{Mean Reports and Individual Draws}
For stochastic decoding, let $X$ and $Y$ be the reports elicited separately for $s$ and $T(s)$, with means $\mu_X$ and $\mu_Y$. Two distinct quantities are
\begin{align}
 A_\mu(s)&=|\mu_X+\mu_Y-1|,\label{eq:means}\\
 A_{\rm draw}^{\gamma}(s)&=\mathbb E_{(X,Y)\sim\gamma}|X+Y-1|.\label{eq:draws}
\end{align}
The joint law $\gamma$ specifies how draws are paired. Our individual-draw analysis pairs draws independently, using the product of the two separately elicited distributions. By convexity of absolute value,
\begin{equation}
 A_\mu(s)\leq A_{\rm draw}^{\gamma}(s).
 \label{eq:jensen}
\end{equation}
Complementary means can therefore coexist with nonzero individual-draw discrepancy. For example, two independent reports each taking $0.2$ or $0.8$ with equal probability have $A_\mu=0$ and $A_{\rm draw}=0.3$.

With $n$ draws per orientation, we estimate these quantities by
\begin{align}
 \widehat A_\mu&=|\bar X+\bar Y-1|,\\
 \widehat A_{\rm draw}&=\frac1{n^2}\sum_{i=1}^{n}\sum_{j=1}^{n}|X_i+Y_j-1|.
\end{align}
The $n^2$ cross-products are dependent functions of $2n$ draws, not additional independent samples, and the plug-in estimate of $A_\mu$ is biased upward. Equal mean reports are also weaker than distributional complement consistency, which requires the law of $Y$ to equal the law of $1-X$.

\subsection{Synchronous Feedback and Factual Loss}
Let $\mathcal N_i$ be the ordered neighbors of agent $i$ on a fixed graph. A synchronous simulation updates
\begin{equation}
 x_i(t+1)=F_\theta(c,x_i(t),(x_j(t))_{j\in\mathcal N_i},E,a).
 \label{eq:synchronous}
\end{equation}
All requests at round $t+1$ read the complete previous-round state. A second population $y$ uses $\bar c$ and starts at $y_i(0)=1-x_i(0)$. Map it back to original coordinates with $z_i(t)=1-y_i(t)$, and measure
\begin{equation}
 d(t)=\frac1N\sum_{i=1}^N|x_i(t)-z_i(t)|.
 \label{eq:trajectory}
\end{equation}
Here $N$ is the number of agents and $t$ is the round. Initially $d(0)=0$. If every local update satisfies Equation~\ref{eq:consistent}, induction gives $y_i(t)=1-x_i(t)$ for every agent and round: complementary states at one round create complementary requests, hence complementary outputs at the next. This requires the same graph, the same evidence, and execution rules that do not depend on the wording. For stochastic updates the random draws must also be coupled compatibly; matching seed integers is not enough.

Once paths diverge, their later requests are no longer exact complements, so later $d(t)$ measures how far the paired trajectories have separated rather than a one-step error. Whether local inconsistency produces growing divergence or changes an intervention ranking is an empirical question, addressed by the feedback and network studies.

Factual performance uses Brier loss
\begin{equation}
 B_\ell(q)=(q-\ell)^2,\qquad B_{1-\ell}(1-q)=B_\ell(q).
 \label{eq:brier}
\end{equation}
For orientation $o$, $L_o(t)$ averages the agents' losses in that orientation. We report $L_0(t)$, $L_1(t)$, and their equal-weight mean, together with the mean over update rounds. Lower loss is better. Consistency is not accuracy: a constant report of $0.5$ is perfectly consistent and has loss 0.25 against either binary label.

\section{Experimental Design}
\subsection{Simulation, Preserved States, and Real Claims}
The simulation study evaluates whether evidence-targeting conclusions change with the update schedule. Sixteen fictional campus claims form eight matched true and false pairs. Each is simulated with twelve world seeds, twelve agents, six rounds, three schedules, and three evidence policies, giving 1,728 cells and 124,416 agent updates. The graph is a cycle augmented with independent chords of probability $0.16$. At round two, the official source is delivered once to the three highest-degree agents, three uniformly selected agents, or no agents. Later requests retain the numerical state and peer reports, rather than the source text.

Synchronous updates read the beginning-of-round snapshot. Fixed sequential updates follow one shuffled agent order throughout a run; randomized sequential updates reshuffle that order each round. Sequential agents can observe already-updated neighbors. The random draws for the graph, initial states, recipients, update order and responses are generated separately and matched across comparisons. Let $L_{j,w,\sigma,\pi}$ be the final population Brier loss for scenario $j$, world seed $w$, schedule $\sigma$, and targeting policy $\pi$. The targeting contrast and primary schedule interaction are

\begin{align}
\Delta_\sigma &= \bigl\langle L_{j,w,\sigma,\mathrm{rnd}} - L_{j,w,\sigma,\mathrm{deg}} \bigr\rangle_{j,w}, \label{eq:delta-sigma}\\
I &= \Delta_{\mathrm{sync}} - \Delta_{\mathrm{rseq}}. \label{eq:interaction}
\end{align}

Here rnd and deg denote random and degree targeting, and sync and rseq the synchronous and randomized sequential schedules.

Positive $\Delta_\sigma$ favors degree targeting. The brackets average matched scenarios and seeds. $I$ measures schedule dependence of that targeting contrast, not complement consistency.

The local criterion has a direct connection to these population outcomes. In a deterministic simulator, complement-consistent local updates preserve complementary trajectories when the graph, source exposure, recipient selection, and execution order do not depend on the wording. The proof proceeds by rounds for synchronous updates and agent by agent for sequential updates. Applying the Brier identity at every final agent state gives
\begin{equation}
 \begin{aligned}
 L^{T}_{j,w,\sigma,\pi}=L_{j,w,\sigma,\pi},\\
 \Delta^T_\sigma&=\Delta_\sigma,\qquad I^T=I.
 \end{aligned}
 \label{eq:policy-invariance}
\end{equation}
Here the superscript $T$ denotes the fully complemented simulation, rather than matrix transposition. This is a sufficiency result: local consistency preserves the policy comparison. The schedule result and the local condition are independent: a small interaction is compatible with large local discrepancies, and large discrepancies are compatible with an unchanged policy ranking.

The preserved-state study takes the first post-delivery update of the three informed agents, at round three, under synchronous and randomized sequential updating and both targeting policies. This gives 2,304 recipient states in 768 matched blocks. Each source-absent response is paired with three additional conditions crossing claim negation and restoration of the exact original source. Own and peer probabilities are transformed together; modified outputs are evaluated for one update and are not propagated.

The real-claim study uses 48 screened AVeriTeC development claims, balanced between Supported and Refuted, from 25 fact-checking organizations \citep{schlichtkrull2023}. We selected standalone claims with linked textual evidence, at most three per organization, before collecting any model responses. Each claim crosses two initial positions, five opposition levels in steps of $25\%$, absent/visible source evidence, original/negated wording, private/public audience, and two mirrored peer orders. Four peers report $0.2$ or $0.8$. The design contains 7,680 response records; the two peer-order reports are averaged before nonlinear outcomes are computed, leaving 3,840 observations and 1,920 complement pairs. Their absolute discrepancy is therefore the absolute discrepancy of order-averaged reports, rather than an average of per-order absolute discrepancies.

These three studies use Qwen at temperature 0.2. The real-claim negations are constructed linguistic pairs, some of which differ in tense or modality; the controlled benchmark below provides a separate test with explicitly exhaustive alternatives. Source restoration changes information access, whereas the evidence-presentation controls below preserve the factual information.

\subsection{Attraction, Accuracy, and Evidence Access}
Let $m_r=k^{-1}\sum_{j=1}^k r_j$ be the displayed peer mean. Peer attraction and one-step accuracy deterioration are
\begin{align}
 C(s;q)&=|p-m_r|-|q-m_r|,\\
 D_B(s;q,\ell)&=B_\ell(q)-B_\ell(p).
 \label{eq:attraction-harm}
\end{align}
Positive $C$ means movement toward the peer mean; positive $D_B$ means a higher Brier loss. The joint event $C>0,D_B>0$ identifies attraction accompanied by a loss of accuracy. Attraction alone can improve or worsen accuracy, depending on the peer information.

For the preserved states, write $q_{o,e}$ for orientation $o\in\{0,1\}$ and source condition $e\in\{0,1\}$, where $e=1$ restores the source. The source-absent mean signed discrepancy is $A_0=\langle q_{0,0}+q_{1,0}-1\rangle$. The average one-step benefit of source restoration is
\begin{align}
R_{\mathrm{src}}=\tfrac12\big\langle &B_\ell(q_{0,0})+B_{1-\ell}(q_{1,0})\nonumber\\
 &-B_\ell(q_{0,1})-B_{1-\ell}(q_{1,1})\big\rangle.
 \label{eq:source-benefit}
\end{align}
The averages first combine the three recipients in each matched block. Positive $R_{\mathrm{src}}$ means lower average loss after restoration.

The real-claim source test asks a different question: does evidence protect an initially correct position against misleading peers? Truth-directed probability is $t_\ell(q)=\ell q+(1-\ell)(1-q)$. For claim $j$, let $\tau_{j,e}(d)$ average this quantity over wording and audience after peer-order averaging, at opposition level $d$ and an initially truth-aligned position. Define
\begin{align}
 D_{j,e}&=\tau_{j,e}(0)-\tau_{j,e}(100),\\
 P&=\frac1J\sum_{j=1}^{J}(D_{j,0}-D_{j,1}),\qquad J=48.
 \label{eq:source-protection}
\end{align}
Positive $P$ means evidence reduces the endpoint decline caused by replacing unanimously agreeing peers with unanimously misleading peers. $R_{\rm src}$ is an average one-step accuracy gain, whereas $P$ is an evidence-by-opposition interaction.

Persuasion direction is defined in displayed-claim coordinates. Let $q(d)$ denote the report at opposition $d$, with other factors matched. The endpoint movements are
\begin{align}
 S_{+\to-}&=\langle q(0)-q(100)\rangle_{p>1/2},\\
 S_{-\to+}&=\langle q(100)-q(0)\rangle_{p<1/2}.
 \label{eq:persuasion}
\end{align}
These average first within claim over source, wording, and audience, then equally over claims. They measure probability movement rather than a boundary-flip rate. Audience contrasts compare public and private instructions for the same claim and remaining factors.

Intervals use 10,000 bootstrap resamples. Network and preserved-state intervals resample matched scenario pairs crossed with world seeds. Real-claim intervals resample claims within label stratum. Network schedule bands are simultaneous 95\% bands; the practical decision margin, set before the network analysis, is $0.02$ Brier units. The two primary endpoints in each paired study use $97.5\%$ intervals for within-study Bonferroni adjustment; secondary intervals are pointwise $95\%$.

\subsection{Materials and Models}
The controlled benchmark contains 24 worlds, four in each of six structural families: exclusive color, binary switch, integer parity, threshold partition, finite-set membership, and strict event order. Every world defines two exhaustive, mutually exclusive outcomes at a fixed referent and time. In half of the worlds the original description is true. Each orientation has a base sentence and one same-polarity paraphrase. The main test describes both outcomes affirmatively (for example, ``the token is red'' and ``the token is blue''); a secondary test replaces the second description with an explicit negation of the first.

With evidence visible, the agent sees an audit record stating which outcome occurred. With evidence absent, it sees only the definition of the two possible outcomes, so its accuracy in that condition is not interpreted as factual reasoning. Both orientations receive the same evidence. We checked programmatically that the two descriptions of every world are exhaustive and mutually exclusive; the wording was not validated by human raters.

\begin{table}[t]
\centering
\small
\setlength{\tabcolsep}{4pt}
\begin{tabular}{lll}
\toprule
Model & Quantization & Instruction in \\
\midrule
Qwen3-4B-Inst.-2507 & Q4\_K\_M  & System \\
Llama-3.2-3B-Inst.  & Q4\_K\_M  & System \\
Gemma-3-4B-IT       & QAT Q4\_0 & User message \\
\bottomrule
\end{tabular}
\caption{Tested models. They differ in size, quantization and prompt format, so differences between them are not attributable to architecture alone. The network, preserved-state and real-claim studies use Qwen at temperature 0.2; the controlled studies use greedy decoding except repeated sampling.}
\label{tab:protocols}
\end{table}

All models run locally through llama.cpp (Table~\ref{tab:protocols}) and must return a single probability in $[0,1]$. Qwen and Llama receive the instruction as a system message. Gemma's chat format has no system role, so the same instruction opens the user message. Llama's chat template inserts the current date, which we fix so that all compared Llama requests are identical. Because the models also differ in size, quantization and prompt format, comparisons are between model configurations rather than architectures. The controlled studies use greedy decoding except the repeated-sampling study, which uses temperature $0.2$ with a separate seed for each draw. Appendix A gives the prompts and one filled-in complementary pair; Appendix D gives the decoding settings.

\begin{table*}[t]
\centering
\small
\begin{tabular}{@{}l p{0.55\textwidth} r@{}}
\toprule
Study & Design & Scale \\
\midrule
Simulation sweep & 16 claims; 12 seeds; 3 schedules; 3 policies; 12 agents; 6 rounds & 124,416 agent updates \\
Preserved states & 2,304 agent states; 4 wording and source conditions (original response reused) & 6,912 new cases \\
Real claims & 48 claims; 2 priors; 5 opposition levels; 2 evidence conditions; 2 wordings; 2 audiences; 2 peer orders & 7,680 cases \\
\midrule
Local consistency & 3 models; 24 worlds; 2 priors; 3 opposition levels; 2 evidence conditions; 2 orientations; 2 wordings & 3,456 \\
Explicit negation & 3 models; same worlds and factors; only the new second orientation is generated & 1,728 \\
Repeated sampling & 3 models; 6 worlds; 80 draws per request and orientation; both evidence conditions & 5,760 \\
Prompt controls & 3 models; 6 worlds; 4 variants; prose and order only with evidence & 216 \\
Paired propagation & 3 models; 6 worlds; 8 agents; 2 initial placements; 5 synchronous rounds & 5,760 \\
Closed-loop repair & 3 models; 6 worlds; 1 placement; reminder, greedy averaging and symmetrization, including a rerun of the Llama baseline after fixing the template date & 12,480 \\
Peer sensitivity & 3 models; 6 worlds; 2 priors; 3 peer counts; 3 report pairs; 3 opposition levels; both evidence conditions and orientations & 3,888 \\
\midrule
Controlled studies, total & New model calls & 33,288 \\
\bottomrule
\end{tabular}
\caption{Design of all studies. The first three rows are the Qwen studies and count agent updates or cases; the remaining rows are the controlled studies and count new model calls, 33,288 in total. Counts are not independent sample sizes. The first two controlled studies use all 24 worlds; the others use six.}
\label{tab:design}
\end{table*}

\subsection{Controls, Feedback, and Repairs}
Table~\ref{tab:design} summarizes the studies. The local tests cross priors of $0.25$ and $0.75$, peer opposition of 0, 50 and 100\%, absent and visible evidence, both orientations, and both wordings. Four ordered peers report $0.2$ or $0.8$, with opposing reports first, and the audience is private. Each numeric input is accompanied by a fixed sentence stating the probability.

The follow-up studies use six of the 24 worlds, one per family. Repeated sampling draws 80 reports per request and orientation, with a fixed world-specific prior, 50\% opposition, and base wording. Four prompt controls test neutral task wording, a reminder that the two outcomes are complementary, evidence rendered as prose, and reordered evidence items; the last two apply only with evidence.

Propagation uses eight agents on a bidirectional radius-two ring, so each agent has four neighbors, ordered by ID, and runs five synchronous rounds. Two placements assign the same initial values $[0.20,0.25,0.35,0.45,0.55,0.65,0.75,0.80]$ to agents in different orders. With evidence visible, every round supplies the same evidence; otherwise none is given. Reports become the next own and peer states without smoothing or memory.

Repair uses one placement and compares four arms: the unchanged baseline, the complement reminder, the average of two greedy calls in the same orientation, and symmetrization, which averages each report with the complement of its paired report (Equation~\ref{eq:repair}). Baseline, reminder and averaging populations evolve separately in each orientation. Symmetrization keeps one population and derives its complementary display from the same recorded pair. Because repaired states change later requests, every round is generated afresh rather than corrected after the fact.

A peer study crosses peer counts of 2, 4 and 8 with symmetric report pairs $0.1/0.9$, $0.2/0.8$ and $0.4/0.6$, both priors, all three opposition levels, and both evidence conditions and orientations, using base wording and the six-world subset. The four-peer condition is regenerated within this study. Changing the count also changes prompt length, and changing report strength changes the distance to the own prior, so neither factor is isolated.

\subsection{Weighting and Uncertainty}
We average discrepancies within paired conditions, then weight worlds, families and factors equally. Agents, rounds, shared baselines and related templates are not independent, and the fixed materials do not support confidence intervals over unseen worlds, graphs, people, or models. Stochastic results carry pointwise Monte Carlo bounds that treat draws as independent; these are not simultaneous or population intervals.

The local-consistency and negation studies cover all 24 worlds; the sampling, prompt-control, propagation, repair and peer studies use six worlds, one per family, and were designed after the first results. Together these studies comprise 33,288 new model calls. Absolute levels on the six-world subset are not comparable with the 24-world means in Figure~\ref{fig:local}: for Qwen with visible evidence the discrepancy is $0.17$ on all 24 worlds but about 0.3 on the subset.

\section{Results}
\subsection{Scheduling and Targeting Measure a Different Property}
\begin{table}[t]
\centering
\begin{tabular}{@{}lrr@{}}
\toprule
Schedule & $\Delta_\sigma$ & Simultaneous $95\%$ interval\\
\midrule
Synchronous & 0.0063 & $[-0.0095,0.0220]$\\
Fixed sequential & 0.0109 & $[-0.0048,0.0267]$\\
Randomized seq. & 0.0048 & $[-0.0109,0.0206]$\\
\bottomrule
\end{tabular}
\caption{Degree-versus-random targeting contrast in final Brier loss, with crossed scenario-pair/world-seed bootstrap bands. Positive values favor degree targeting. The primary schedule interaction is reported separately in the text.}\label{tab:network}
\end{table}

\begin{figure*}[t]\centering
\includegraphics[width=\textwidth]{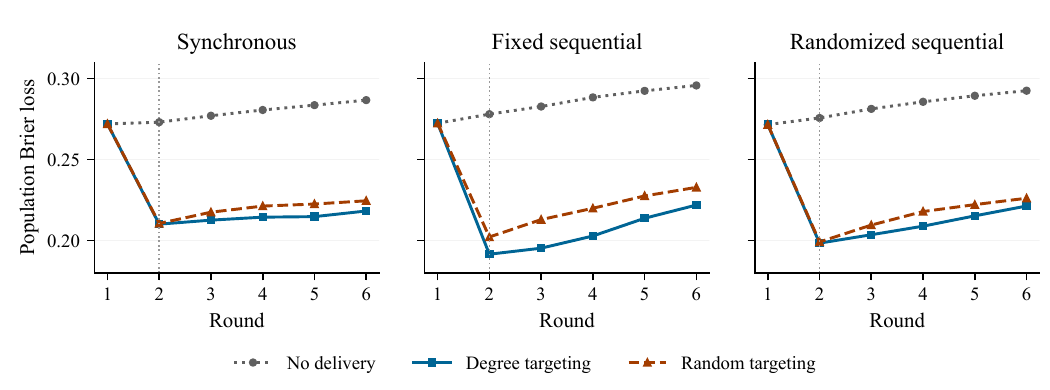}
\caption{Population Brier loss over six rounds for the three evidence policies under each update schedule, averaged over 16 scenarios and 12 world seeds. The dotted line marks the one-time source delivery at round two; later requests carry probabilities but not the source text.}\label{fig:network}\end{figure*}

The degree-versus-random targeting contrasts are positive point estimates under all three schedules (Table~\ref{tab:network}). The primary interaction is $I=0.0015$, with $95\%$ interval $[-0.0093,0.0132]$. The intervals cross zero: the experiment does not establish a targeting advantage or a schedule interaction. The decision rule set in advance for the network study does not identify equivalence from this result.

Both delivery policies produce substantially lower final Brier loss than no delivery at the point-estimate level. Figure~\ref{fig:network} shows the immediate source response and subsequent drift after the source text is no longer supplied. Scheduling also changes final state variance: the degree-targeted synchronous and fixed-sequential means are $0.0049$ and $0.0024$. These outcomes concern targeting and dispersion. Scheduling also changes the final variance of agent states: under degree targeting it is $0.0049$ for synchronous and $0.0024$ for fixed sequential updating.

\subsection{Complement Tests on Simulation States and Real Claims}
\begin{table*}[t]
\centering
\begin{tabular}{@{}lrrl@{}}
\toprule
Probe / estimand & Estimate & Interval & Level\\
\midrule
Preserved states: signed $A_0$ & 0.1846 & $[0.1553,0.2152]$ & 97.5\%\\
Preserved states: Brier gain $R_{\mathrm{src}}$ & 0.0233 & $[-0.0450,0.0763]$ & 97.5\%\\
Preserved states: absolute discrepancy & 0.1899 & $[0.1650,0.2156]$ & 95\%\\
Real claims: signed discrepancy & 0.1867 & $[0.1281,0.2449]$ & 97.5\%\\
Real claims: protection $P$ & 0.0250 & $[-0.0501,0.0994]$ & 97.5\%\\
Real claims: absolute discrepancy & 0.3259 & $[0.2929, 0.3622]$ & 95\%\\
\bottomrule
\end{tabular}
\caption{Paired-probe results on simulation states and real-claim requests. Preserved-state diagnostics average three recipients within each matched block. Real-claim outcomes average peer orders before nonlinear calculations, then weight claims equally. The primary 97.5\% intervals adjust for two endpoints per study; the two absolute-discrepancy intervals are secondary and pointwise. These are separate estimands, not comparable standardized effect sizes.}\label{tab:original-pairs}
\end{table*}

The preserved recipient states connect the test to actual simulation inputs. Without source text, their mean signed discrepancy is $0.1846$ and mean absolute discrepancy is $0.1899$ (Table~\ref{tab:original-pairs}). The responses on complementary requests therefore do not sum to one on average. Restoring the source changes the signed mean to $-0.1138$, with pointwise $95\%$ interval $[-0.2278,-0.0208]$. The direction changes with the information available in the request.

The real-claim requests have mean signed discrepancy $0.1867$ and mean absolute discrepancy $0.3259$ after peer-order averaging. Their median absolute discrepancy is $0.2500$. The two means differ because positive and negative discrepancies cancel in the signed average. Restoring consistency would require changing each report by $|\delta|/2$, which averages 0.163. The label strata further separate them: Supported claims have signed and absolute means of $0.3488$ and $0.3647$, whereas Refuted claims have $0.0246$ and $0.2872$.

These results concern single stochastic reports from Qwen, on simulation states and on negations we wrote for real claims. The following controlled benchmark tests the same update criterion with explicit binary partitions, matched same-polarity paraphrases, repeated draws, and additional models.

\subsection{Attraction, Accuracy, and Consistency Separate Empirically}
\begin{table*}[t]
\centering
\begin{tabular}{@{}lrr@{}}
\toprule
Secondary contrast & Estimate & Pointwise $95\%$ interval\\
\midrule
Preserved states: source contrast in $C$ & 0.1390 & $[0.1240,0.1559]$\\
Preserved states: source contrast in $\Pr(C>0,D_B>0)$ & 0.2947 & $[0.2639,0.3236]$\\
Real claims: $S_{+\to-}-S_{-\to+}$ & -0.1555 & $[-0.1940,-0.1162]$\\
Real claims: public--private $C$ & 0.0116 & $[0.0075,0.0156]$\\
Real claims: public--private $D_B$ & -0.0027 & $[-0.0073,0.0020]$\\
\bottomrule
\end{tabular}
\caption{Attraction, factual loss, and persuasion are measured separately. Source contrasts are absent minus restored; audience contrasts are public minus private. The joint-event proportion is on a 0--1 scale. Preserved-state intervals use crossed scenario-pair/world-seed resampling; real-claim intervals resample claims within label.}\label{tab:behavior}
\end{table*}

Source restoration changes mean peer attraction from $0.0794$ to $-0.0596$ and reduces the proportion of updates with $C>0,D_B>0$ from $31.47\%$ to $2.00\%$. The remaining joint events have larger conditional mean accuracy deterioration: $0.7825$ with source restoration versus $0.1807$ without it. The event sets differ, so this compares the severity of the events that remain, rather than a matched increase in individual harm. The primary average Brier-gain estimate $R_{\mathrm{src}}=0.0233$ has an interval spanning both signs. Reduced attraction and fewer harmful-attraction events therefore do not establish an average accuracy gain.

The real-claim evidence-protection estimate $P=0.0250$ also has an interval spanning zero. It measures resistance of initially correct positions to misleading peers, whereas the one-step source test averages accuracy changes across positions. Figure~\ref{fig:original-dose} separates correction of initially incorrect positions from protection of initially correct ones.

Movement from negative to affirmative endpoints is $0.3830$, compared with $0.2275$ in the reverse direction. This extends an established persuasion-asymmetry question to the present protocol \citep{mehdizadeh2025}. Telling agents that their answers will be shared increases peer attraction by $0.0116$, while the accuracy-deterioration contrast is $-0.0027$ with an interval crossing zero (Table~\ref{tab:behavior}). Numerical social movement, factual accuracy, and complement consistency are distinct empirical outcomes.

\subsection{Complement Discrepancy and Paraphrase Sensitivity}
\begin{figure*}[t]\centering
\includegraphics[width=\textwidth]{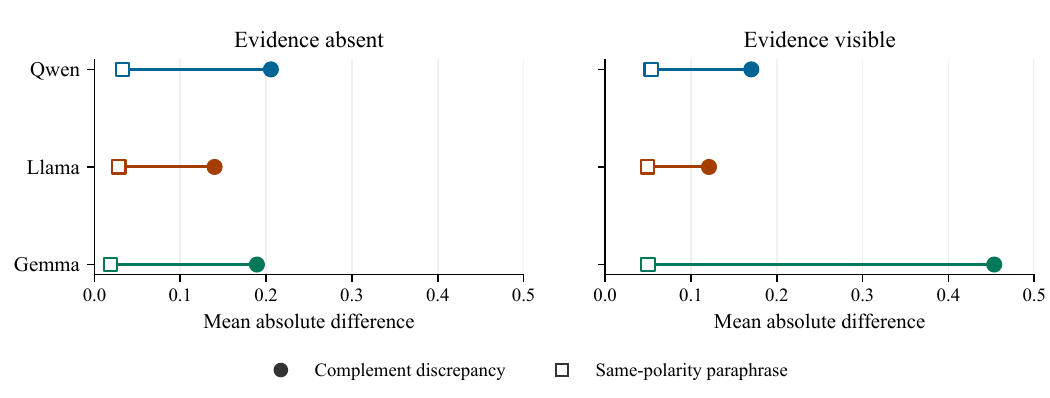}
\caption{Mean absolute complement discrepancy (filled circles) and same-polarity paraphrase sensitivity (open squares) on the 24-world benchmark, by model and evidence condition. Differences are computed within pairs before worlds and families are weighted equally. Complement discrepancy is larger in all six cases.}\label{fig:local}\end{figure*}

Figure~\ref{fig:local} compares absolute complement discrepancy with matched paraphrase sensitivity. Complement discrepancy is larger for every model and evidence condition. Not every request fails: with visible evidence, 215 of 288 Qwen pairs and 210 of 288 Llama pairs are exactly consistent. Signed averages can hide opposing errors, particularly for Qwen with evidence.

Evidence has no uniform effect on consistency. On all 24 worlds, Qwen's and Llama's discrepancies decrease when evidence is visible, whereas Gemma's increases; on the six-world subset used in the follow-up studies, Qwen's is higher with evidence (Figures~\ref{fig:appendix-S5_sampling_estimands} and~\ref{fig:appendix-M5_peer_grid}). The discrepancy measures consistency, not accuracy, so it does not rank the models by factual performance. Results by family appear in Appendix C.

Explicit negation increases absolute discrepancy for Qwen and Llama in both evidence conditions but decreases it for Gemma. Difficulty with negation alone therefore cannot explain the results obtained with affirmative alternatives.

\subsection{Repetition and Prompt Controls}
The repeated-sampling study separates the two quantities in Equations~\ref{eq:means} and~\ref{eq:draws}. Mean-level discrepancies persist, so averaging does not remove them. For Llama with visible evidence, the mean-level discrepancy is $0.168$, against $0.184$ for independently paired draws and $0.200$ under greedy decoding. For Qwen with visible evidence, all 80 draws of each request are identical, so the two estimates coincide.

Prompt effects are mixed. Rendering Qwen's evidence as prose reduces its discrepancy but worsens visible Brier loss in the complementary orientation. The complement reminder improves Gemma's consistency and visible loss, leaves Llama's visible consistency unchanged while slightly worsening its loss, and increases Qwen's discrepancy without evidence by about $0.1$. Reordering evidence items can change reports without adding information. These controls show sensitivity to instructions and presentation but do not identify its cause. Complete results appear in Appendix C.

\subsection{Paired Feedback Trajectories}
\begin{figure*}[t]\centering
\includegraphics[width=\textwidth]{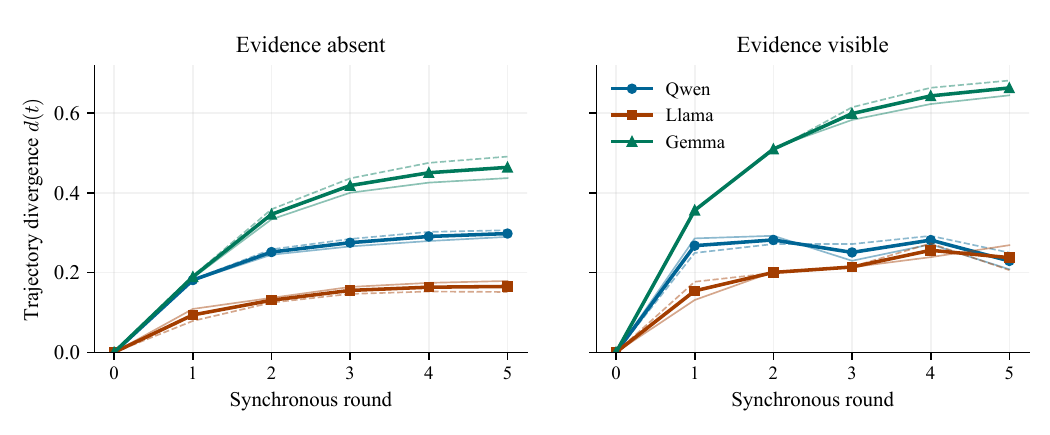}
\caption{Divergence $d(t)$ between paired populations that start from complementary states, over five synchronous rounds on six worlds. Thick lines average the two initial placements, which are shown as thin solid and dashed lines. With visible evidence, Qwen's divergence stays roughly flat after the first round, while Llama’s and Gemma’s grow.}\label{fig:feedback}\end{figure*}

All paired populations start at zero divergence. Figure~\ref{fig:feedback} shows positive final divergence for every model and evidence condition, with different temporal patterns. With visible evidence, Qwen's divergence stays roughly flat (about 0.23 to 0.28), and some world and placement combinations finish exactly consistent, whereas Llama's and Gemma's divergence grows. Wording dependence therefore enters the feedback process, but it is not always amplified.

As a check on the simulator, an agent that returns the average of its own state and its neighbors' mean keeps the paired paths complementary to within rounding error.

\subsection{Sensitivity to Peer Count and Reports}
All 54 combinations of model, evidence condition, peer count and report pair have positive mean discrepancy, although 624 of the 1,944 underlying pairs are exactly consistent. The inconsistency is therefore not specific to the four-peer setting of the main tests, but its size varies, and the effects of count and report strength differ across models. The full grid appears in Appendix C.

Consistency and accuracy can move in opposite directions. With $0.2/0.8$ reports, increasing Llama's peers from four to eight reduces visible discrepancy by $0.042$ but increases Brier loss averaged over orientations by $0.008$. All peer groups here are homogeneous and present opposing reports first, so heterogeneous groups remain untested.

\section{Averaging, Symmetrization, and Utility}

\begin{figure*}[t]
\centering
\includegraphics[width=0.92\textwidth]{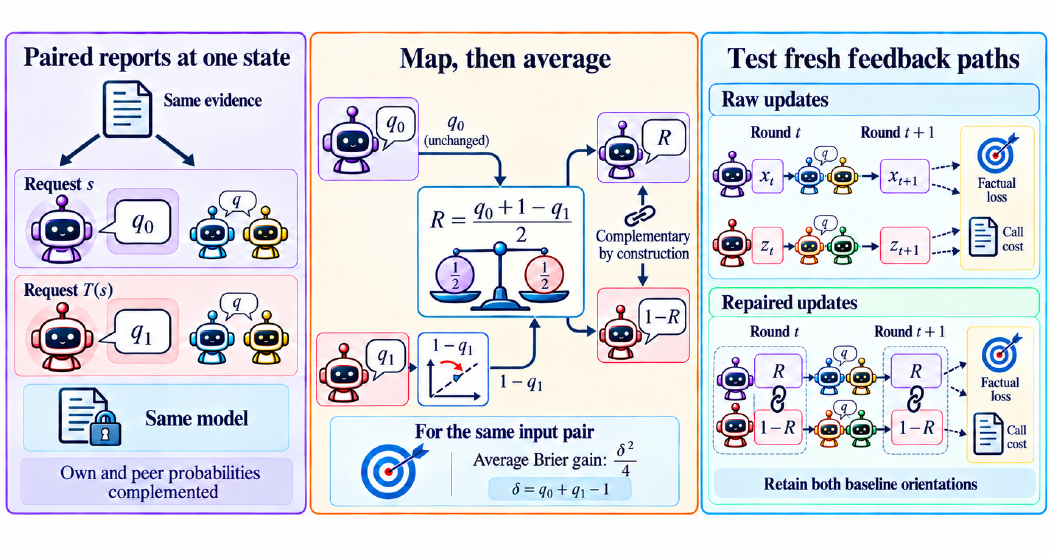}
\caption{Symmetrization. Left: the same state is described in both
orientations and the model answers each request in a separate call, giving
$q_0$ and $q_1$. Middle: $q_1$ is mapped back to the original orientation as
$1-q_1$ and averaged with $q_0$ to give $R$. The repaired pair $(R,1-R)$ is
complementary by construction, and for this fixed pair the average Brier loss
falls by $\delta^2/4$. Right: repaired reports become the next round's states
and are compared with the unrepaired populations on factual loss and call
cost. The one-step identity does not determine the outcome of this
closed-loop comparison.}
\label{fig:repair-mech}
\end{figure*}

\subsection{What the Algebra Guarantees}
For independent repeated reports $Q_1,\ldots,Q_m$ at the same request, the average $\bar Q_m=m^{-1}\sum_j Q_j$ preserves the expected report and has variance $\operatorname{Var}(Q_1)/m$ when draws are identically distributed with finite variance. With correlated draws, covariance terms remain. Averaging therefore need not remove the mean-level discrepancy in Equation~\ref{eq:means}, and identical greedy calls add nothing.

For complementary reports on the same state, $q_0=F_\theta(s)$ and $q_1=F_\theta(T(s))$, define
\begin{equation}
 R(s)=\frac{q_0+1-q_1}{2}.
 \label{eq:repair}
\end{equation}
Both $q_0$ and $1-q_1$ are probabilities in the original orientation, so their average lies in $[0,1]$. Under an involutive $T$ and fixed deterministic evaluations,
\begin{equation}
 R(T(s))=\frac{q_1+1-q_0}{2}=1-R(s).
 \label{eq:repair-proof}
\end{equation}
Swapping the two components therefore produces an exactly complementary output. In our implementation both components are recorded once and the complementary display is taken from that shared pair; recomputing with fresh stochastic draws would not guarantee exact equality.

In displayed-claim coordinates, the repaired pair is $(R(s),1-R(s))$, and both raw reports change by $-\delta(s)/2$. The adjustment is $|\delta(s)|/2$ per coordinate, which is the Euclidean projection of $(q_0,q_1)$ onto the complementary segment $\{(u,1-u):0\leq u\leq1\}$.

\begin{samepage}
Let $u=q_0$, $v=1-q_1$, and $m=(u+v)/2$. Expanding the squared errors gives
\begin{align}
 &\frac{(u-\ell)^2+(v-\ell)^2}{2}-(m-\ell)^2\nonumber\\
 &\quad=\frac{u^2+v^2}{2}-\frac{(u+v)^2}{4}
 =\frac{(u-v)^2}{4}=\frac{\delta(s)^2}{4}.
 \label{eq:brier-proof}
\end{align}
\end{samepage}
Symmetrization therefore cannot increase the Brier loss averaged over orientations for a fixed pair with complementary labels, although it need not improve either orientation individually. This identity compares the same inputs. Once repaired outputs change future states, the baseline and repaired requests differ, and Equation~\ref{eq:brier-proof} says nothing about their closed-loop losses.

\subsection{Closed-Loop Results}
\begin{figure*}[t]\centering
\includegraphics[width=\textwidth]{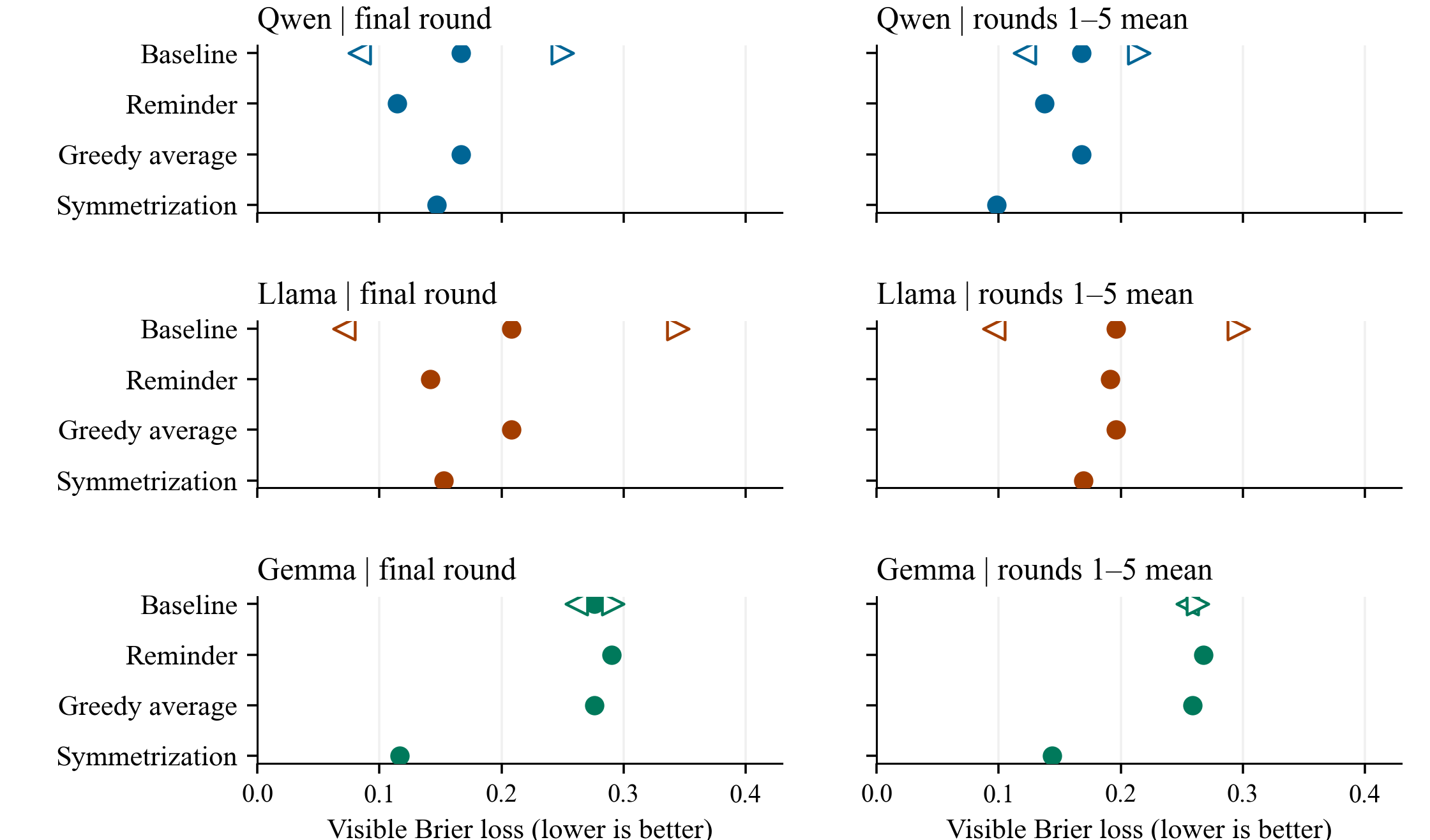}
\caption{Visible-evidence Brier loss after closed-loop feedback on six worlds. Filled circles: loss averaged over the two orientations. Open triangles: the baseline’s original (left) and complementary (right) orientations separately. Symmetrization lowers the averaged loss for all three models but is not always better than the better baseline orientation.}\label{fig:repair}\end{figure*}

Figure~\ref{fig:repair} reports factual loss for each repair arm alongside both baseline orientations. Symmetrization lowers final and round-averaged visible Brier loss, averaged over orientations, for all three models. It does not beat both baseline orientations: Qwen's final loss is worse than its better baseline orientation, and so are Llama's final and round-averaged losses.

The complement reminder improves the orientation-averaged endpoints for Qwen and Llama but worsens Gemma's. For Llama it improves the complementary orientation (0.345 to 0.209) and leaves the original one almost unchanged (0.071 to 0.074). A prompt change that helps locally therefore need not help under feedback. Both orientations for every arm appear in Appendix C.

Each pair of same-orientation greedy calls returns identical numbers, so averaging reproduces the baseline paths exactly. This shows repeatability for the requests visited, not determinism for every possible prompt.

The arms also differ in cost. Baseline and reminder need one call per answer in each orientation, or two for an independently evolving pair of populations; averaging needs two and four. Symmetrization uses two calls for one repaired answer and its complementary display, and its zero divergence holds by construction rather than as an independent result. The arms are not matched for compute.

\section{Discussion}
\paragraph{Local consistency and collective robustness.}
The network sweep and the paired probes measure different properties of the same simulation. The sweep asks how an evidence-targeting contrast depends on execution order; the probes ask how an agent's report depends on how its information is worded. The sufficiency result in Equation~\ref{eq:policy-invariance} connects them: complement-consistent updates preserve a fully transformed policy comparison. The converse is false, so the two kinds of evidence have to be collected separately. The propagation study shows that wording differences can enter feedback without always being amplified.

\paragraph{Consistency and factual utility.}
Symmetrization lowers loss averaged over the two wordings but sometimes does worse than the better original wording, so a repair has to be judged with wording-specific losses and whole-trajectory outcomes.

\paragraph{Implications for evaluation.}
When generated probabilities become simulation states, a consistency test should transform the complete request: proposition, own probability, ordered peer reports, and fixed evidence. It should be paired with same-polarity paraphrase controls, should separate mean-report effects from variation across draws, and should propagate both orientations under identical execution rules. Evaluating a repair also requires factual loss and call cost alongside consistency. The numbers we report are observed responses; reading them as internal beliefs or as human behavior requires further evidence.

\section{Conclusion}
Equivalent complementary descriptions produce incompatible probability reports in the tested agent simulations. The discrepancy appears on states taken from a running simulation and on real fact-checking claims, exceeds matched paraphrase sensitivity in the controlled benchmark, and can persist after reports become later inputs. Symmetrization enforces complementary outputs, but its value under feedback is an empirical question: lower loss averaged over orientations can coexist with higher loss than the better baseline orientation. Simulations that treat generated probabilities as agent states should therefore report consistency and factual performance separately.

\bibliography{references}
% =====================================================================
% appendix.tex  --  \input{appendix} after \bibliography{references}
% Requires in the main preamble (both allowed by the AAAI kit):
%   \usepackage{xcolor}
%   \usepackage[most]{tcolorbox}
% Figure files are the existing ones from the technical appendix:
%   figures/appendix-family, figures/appendix-negation,
%   figures/appendix-sampling, figures/appendix-prompts,
%   figures/appendix-peers, figures/original-dose-response
% =====================================================================

\definecolor{promptbg}{gray}{0.97}
\definecolor{promptframe}{gray}{0.55}
\definecolor{diffcol}{RGB}{176,32,32}
\lstdefinestyle{prompt}{
  basicstyle=\scriptsize\ttfamily,
  breaklines=true, breakatwhitespace=true,
  columns=fullflexible, keepspaces=true, showstringspaces=false,
  escapeinside={(*@}{@*)},
  literate={’}{{\textquoteright}}1 {‘}{{\textquoteleft}}1
           {“}{{\textquotedblleft}}1 {”}{{\textquotedblright}}1
}
\newtcblisting{promptbox}[1]{
  listing only, breakable, enhanced,
  colback=promptbg, colframe=promptframe, boxrule=0.4pt, arc=1pt,
  left=3pt, right=3pt, top=2pt, bottom=2pt,
  fonttitle=\scriptsize\bfseries, coltitle=black, colbacktitle=gray!18,
  title={#1}, listing options={style=prompt,aboveskip=0pt,belowskip=0pt}
}
\newtcblisting{promptfig}[1]{
  listing only, enhanced,
  colback=promptbg, colframe=promptframe, boxrule=0.4pt, arc=1pt,
  left=3pt, right=3pt, top=2pt, bottom=2pt,
  fonttitle=\scriptsize\bfseries, coltitle=black, colbacktitle=gray!18,
  title={#1}, listing options={style=prompt,aboveskip=0pt,belowskip=0pt}
}
\newcommand{\diff}[1]{\textcolor{diffcol}{#1}}

\section{Appendix A: Prompt Templates and Example Inputs}
\label{app:prompts}

The boxes show the messages sent to the models, with protocol version identifiers removed and line breaks adjusted; double-brace slots are filled for each request. Qwen and Llama receive the system instruction and the user message as separate roles. Gemma has no system role, so it receives both as one user turn separated by a blank line. Output is constrained to a JSON object with the single field \texttt{p\_claim\_true}, a number in $[0,1]$.

\begin{promptbox}{System instruction (controlled benchmark and real claims)}
You simulate one participant and must follow output contract.
p_claim_true is the participant’s probability that the exact displayed claim is true.

0.0 means certainly false.
0.5 means maximally uncertain or evenly balanced.
1.0 means certainly true.

Evidence supporting the exact claim should move this value toward 1.
Evidence refuting the exact claim should move this value toward 0.

This value never represents confidence in the response, probability that the evidence is reliable, agreement with a peer, or probability that an alternative claim is true.

Return exactly one JSON object containing only p_claim_true. Do not add any other field. Use only the supplied initial probability, peer reports, source evidence when present, and audience instruction. Do not invent evidence or access hidden facts.

Balanced hypothetical examples, unrelated to the experiment:
- Strong generic support for the displayed claim: {"p_claim_true":0.9}
- Strong generic refutation of the displayed claim: {"p_claim_true":0.1}
- Maximally balanced generic information: {"p_claim_true":0.5}
These examples define output direction only and do not state any experimental answer.
\end{promptbox}

\begin{promptbox}{User message template (controlled benchmark)}
Apply output contract.

Simulate one participant updating a probability for the exact displayed claim. Treat the initial probability as the participant’s current position, not as independent factual evidence. Peer reports are other participants’ numeric positions, not verified evidence. Use the source evidence only when it is displayed. Do not invent evidence, retrieve outside facts, or infer hidden experimental metadata.

Task facts (fixed in every condition):
{{CONTEXT}}

Your response will remain private and will not be shown to the peer group.

Observation:
{{OBSERVATION_JSON}}

{{SOURCE_BLOCK}}

Return JSON only with exactly one field named p_claim_true.
\end{promptbox}

Figure~\ref{fig:prompt-pair} shows how the slots are filled for one complementary pair. Everything outside the observation is identical in the two requests, including the task facts, the evidence block and the audience sentence. When evidence is absent, the evidence block is replaced by the single line \emph{No source evidence is visible in this update.}

\begin{figure*}[t]
\centering
\begin{minipage}[t]{0.49\textwidth}\vspace{0pt}
\begin{promptfig}{Original orientation ($o=0$): observation}
{
  "displayed_claim": "The token in exclusive_color trial 1 is (*@\diff{red}@*) at tick 101.",
  "initial_state": {
    "message": "I currently assign probability (*@\diff{0.25}@*) that the displayed claim is true.",
    "p_claim_true": (*@\diff{0.25}@*)
  },
  "peer_reports_in_display_order": [
    {"agent": "peer-1", ..., "p_claim_true": (*@\diff{0.8}@*)},
    {"agent": "peer-2", ..., "p_claim_true": (*@\diff{0.8}@*)},
    {"agent": "peer-3", ..., "p_claim_true": (*@\diff{0.2}@*)},
    {"agent": "peer-4", ..., "p_claim_true": (*@\diff{0.2}@*)}
  ]
}
\end{promptfig}
\end{minipage}\hfill
\begin{minipage}[t]{0.49\textwidth}\vspace{0pt}
\begin{promptfig}{Complementary orientation ($o=1$): observation}
{
  "displayed_claim": "The token in exclusive_color trial 1 is (*@\diff{blue}@*) at tick 101.",
  "initial_state": {
    "message": "I currently assign probability (*@\diff{0.75}@*) that the displayed claim is true.",
    "p_claim_true": (*@\diff{0.75}@*)
  },
  "peer_reports_in_display_order": [
    {"agent": "peer-1", ..., "p_claim_true": (*@\diff{0.2}@*)},
    {"agent": "peer-2", ..., "p_claim_true": (*@\diff{0.2}@*)},
    {"agent": "peer-3", ..., "p_claim_true": (*@\diff{0.8}@*)},
    {"agent": "peer-4", ..., "p_claim_true": (*@\diff{0.8}@*)}
  ]
}
\end{promptfig}
\end{minipage}

\vspace{2pt}
\begin{minipage}[t]{0.49\textwidth}\vspace{0pt}
\begin{promptfig}{Task facts (shared)}
In exclusive_color trial 1, a sealed box at tick 101 contains exactly one token. Its color is exactly one of red and blue. Colors do not change during this trial.
\end{promptfig}
\end{minipage}\hfill
\begin{minipage}[t]{0.49\textwidth}\vspace{0pt}
\begin{promptfig}{Evidence block, visible condition (shared)}
Source evidence is visible:
{
  "facts": {"token_color": "red"},
  "referent": "exclusive_color trial 1",
  "source": "exact audit log for exclusive_color trial 1",
  "time": "tick 101",
  "uncertainty": "The audit is stipulated to be exact and error-free."
}
\end{promptfig}
\end{minipage}
\caption{One complementary pair from the color family with visible evidence, the case drawn in Figure~\ref{fig:overview}. Red marks every value that differs between the two requests. Each peer's \texttt{message} repeats the sentence used for \texttt{initial\_state} with that peer's value and is abbreviated here only. The model's output for each request is $q_0$ and $q_1$, and consistency requires $q_0+q_1=1$.}
\label{fig:prompt-pair}
\end{figure*}

\paragraph{Prompt controls.} Each control is applied on its own. The complement reminder below is appended to the task paragraph and is also one of the repair arms. Neutral wording rephrases the task paragraph without changing its instructions, evidence prose rewrites the evidence block as sentences, and evidence reordering changes the order of the evidence fields without changing their content.

\begin{promptbox}{Complement reminder}
For any two mutually exclusive and exhaustive complementary propositions, their probabilities sum to one. Apply this probability constraint to the displayed claim while respecting the displayed prior, peer reports, and source evidence. Return only the probability of the displayed claim in the required JSON field.
\end{promptbox}

\paragraph{Network simulation and preserved states.} These studies use the same system instruction, reworded for a fictional campus participant. The task paragraph asks the model to update its probability given peer reports and any official timetable or route it receives. There is no task-facts block or audience sentence; delivered evidence is part of the observation.

\paragraph{Real claims.} The real-claim study uses the controlled system instruction and user template with these changes: the task paragraph refers to ``the source question-answer evidence'', there is no task-facts block, the observation holds an AVeriTeC claim (for example, \emph{Schools have resumed in New Brunswick, Canada despite the COVID-19 pandemic}), and peers carry no agent label. The public condition replaces the audience sentence with \emph{Your response will be shared with the peer group after this update.}

\paragraph{Feedback.} At round $t+1$ each agent sees its own probability and its neighbors' reports from the complete round-$t$ state; the evidence is unchanged. The baseline and reminder arms feed back their single report, and the averaging arm feeds back the mean of two identical greedy calls. Symmetrization queries both orientations of the same state, obtains $q_0$ and $q_1$, feeds back $R=(q_0+1-q_1)/2$ in the original orientation, and displays $1-R$ in the complementary one.

\section{Appendix B: Construction of the Controlled Benchmark}
\label{app:materials}

Table~\ref{tab:families} lists the six families. Each world fixes a referent and a time, and its two descriptions are exhaustive and mutually exclusive. Both descriptions of a world concern the same observed fact, so one evidence block verifies either of them. This avoids the asymmetry between existential and universal claims that open-world statements have. The explicit-negation control keeps the original description and replaces the complementary one with its negation, for example \emph{is not red} in place of \emph{is blue}.

\begin{table}[h]
\centering
\small
\setlength{\tabcolsep}{4pt}
\begin{tabular}{lll}
\toprule
Family & Original ($o=0$) & Complement ($o=1$) \\
\midrule
Binary switch & on & off \\
Exclusive color & red & blue \\
Integer parity & even & odd \\
Threshold & below & at or above \\
Set membership & in set & in its complement \\
Event order & A before B & B before A \\
\bottomrule
\end{tabular}
\caption{The six structural families, each with four worlds. Every world defines two exhaustive, mutually exclusive outcomes at a fixed referent and time.}
\label{tab:families}
\end{table}

\section{Appendix C: Supplementary Results}
\label{app:results}

\begin{figure*}[t]\centering
\includegraphics[width=\textwidth]{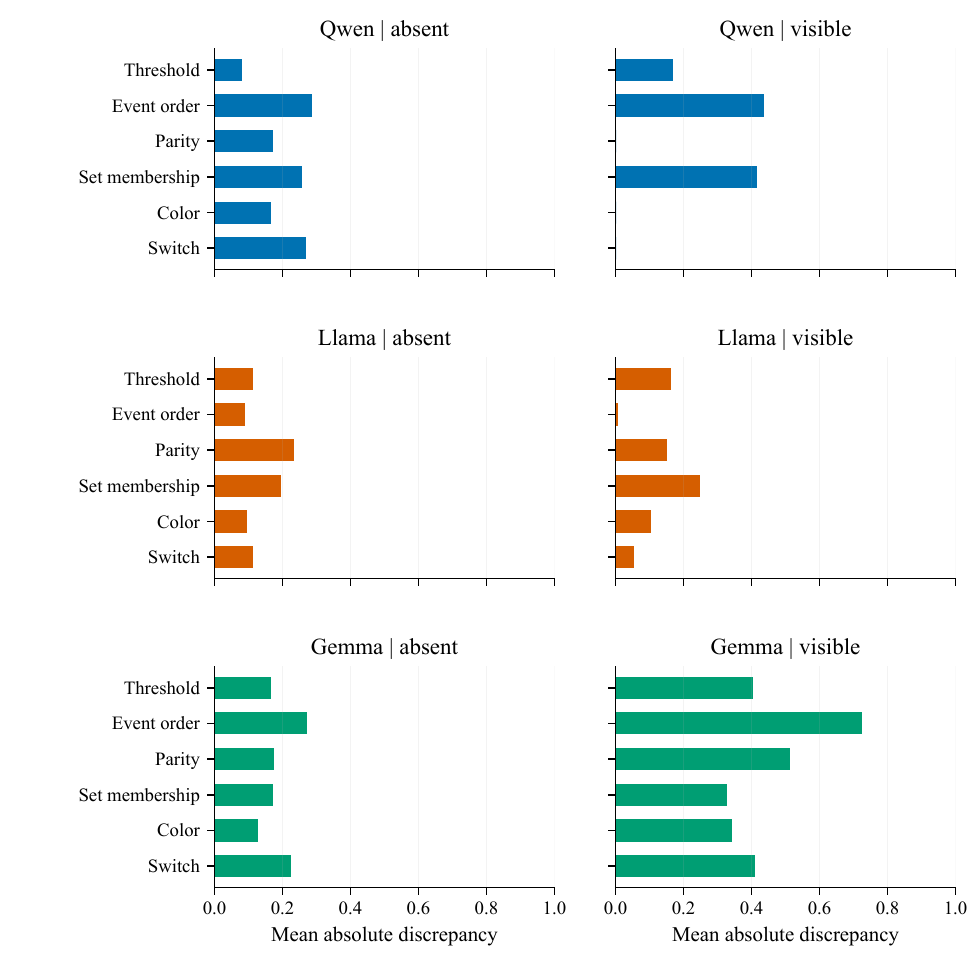}
\caption{Mean absolute complement discrepancy by structural family (four worlds each) with affirmative descriptions, for each model and evidence condition. Bars at zero are measured values.}\label{fig:appendix-S1_family_affirmative}\end{figure*}

\paragraph{Discrepancy by family.} Figure~\ref{fig:appendix-S1_family_affirmative} breaks the controlled discrepancy down by family. No single family produces the effect. Without evidence, all six families show a discrepancy for every model. With evidence, Qwen's discrepancy is concentrated in three families (event order, set membership and threshold), whereas Gemma's large discrepancy is spread across all six.

\begin{figure*}[t]\centering
\includegraphics[width=\textwidth]{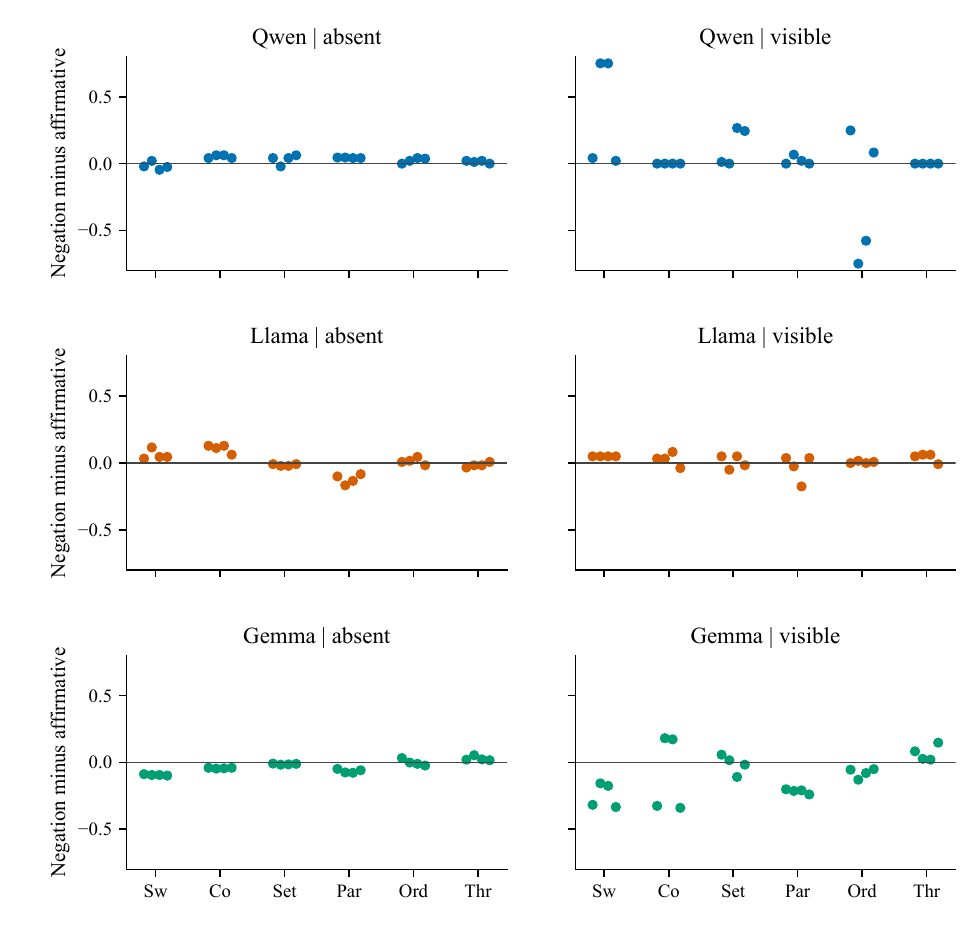}
\caption{Change in absolute discrepancy for each world when the complementary description is an explicit negation instead of an affirmative alternative. Negative values mean the negation reduced the discrepancy. Sw: switch, Co: color, Set: membership, Par: parity, Ord: event order, Thr: threshold.}\label{fig:appendix-S4_negation_changes}\end{figure*}

\paragraph{Explicit negation.} Figure~\ref{fig:appendix-S4_negation_changes} shows, world by world, how the absolute discrepancy changes when the complementary description is an explicit negation. Without evidence, most changes are below 0.15 in magnitude. With evidence, Qwen shows large changes in both directions in a few worlds, Llama's remain small, and Gemma's are mostly negative. Negation therefore leaves the main result intact, and the affirmative descriptions are not systematically easier for the models.

\begin{figure*}[t]\centering
\includegraphics[width=\textwidth]{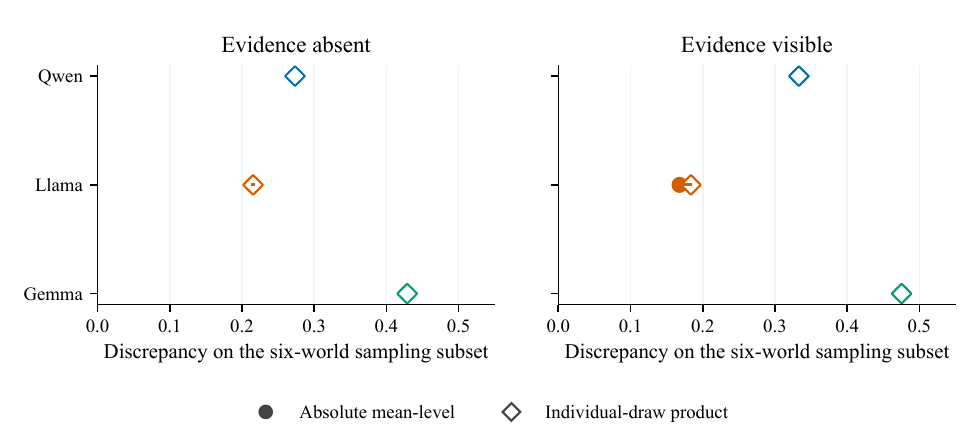}
\caption{Discrepancy of the mean reports (circles) and mean discrepancy of independently paired draws (diamonds), from 80 draws per request on the six-world subset. Where only a diamond is visible, the two coincide.}\label{fig:appendix-S5_sampling_estimands}\end{figure*}

\paragraph{Repeated sampling.} Each request in the six-world subset receives $n=80$ draws at temperature $0.2$. Figure~\ref{fig:appendix-S5_sampling_estimands} compares the discrepancy of the mean reports, $\widehat A_\mu$, with the mean discrepancy of independently paired draws, $\widehat A_{\rm draw}$ (Equations~\ref{eq:means}--\ref{eq:draws}). The two coincide except for Llama with visible evidence, so the discrepancy reflects the mean report rather than sampling noise. Each orientation mean receives a Hoeffding interval with half-width $h=\sqrt{\ln 80/(2n)}$, which covers the true mean with probability at least $0.975$; the pair of intervals jointly covers with probability at least $0.95$. The interval for the signed discrepancy of a pair is $[\,l_X+l_Y-1,\;u_X+u_Y-1\,]$, where $[l_X,u_X]$ and $[l_Y,u_Y]$ are the two mean intervals clipped to $[0,1]$.

\begin{figure*}[t]\centering
\includegraphics[width=\textwidth]{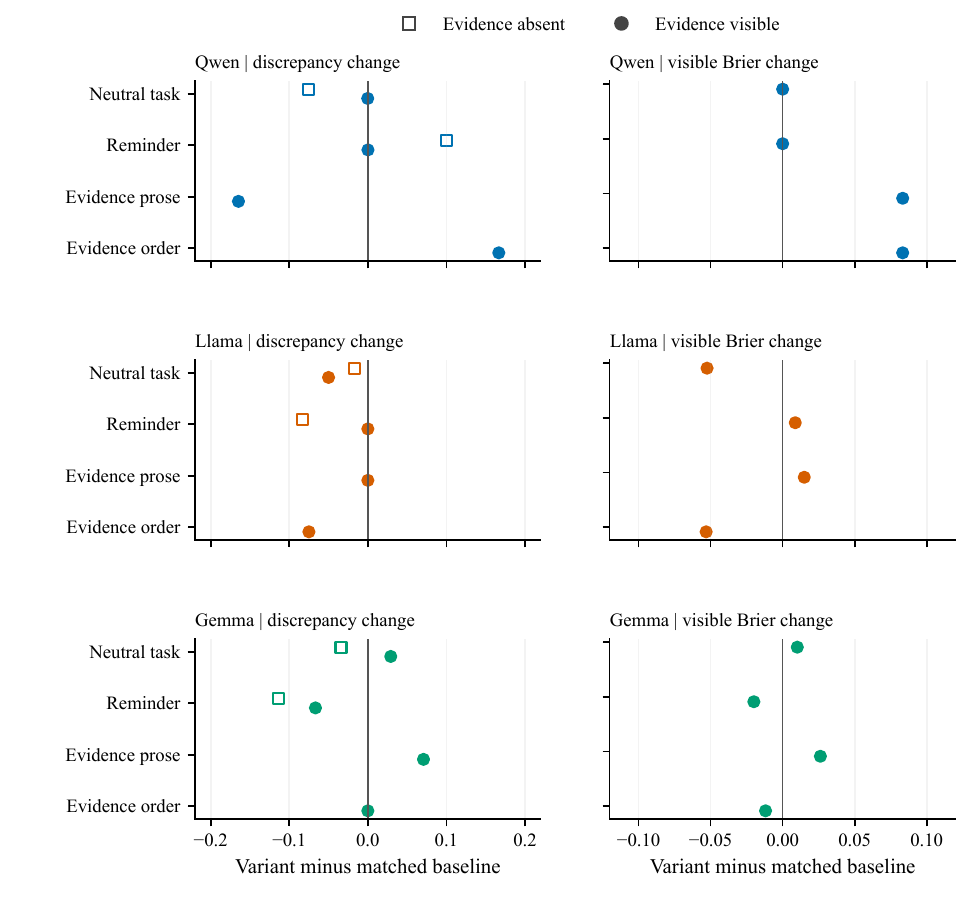}
\caption{Change relative to the matched baseline for each prompt control. Left: absolute complement discrepancy (open squares, evidence absent; filled circles, evidence visible). Right: visible Brier loss averaged over orientations. Evidence prose and reordering apply only with visible evidence.}\label{fig:appendix-S7_prompt_sensitivity}\end{figure*}

\paragraph{Prompt controls.} Figure~\ref{fig:appendix-S7_prompt_sensitivity} gives each control's change in discrepancy and in visible Brier loss relative to the matched baseline. No control reduces the discrepancy for all three models in both evidence conditions, and reductions in discrepancy do not consistently come with lower loss; rendering Qwen's evidence as prose, for example, lowers its discrepancy but raises its loss.

\begin{figure*}[t]\centering
\includegraphics[width=\textwidth]{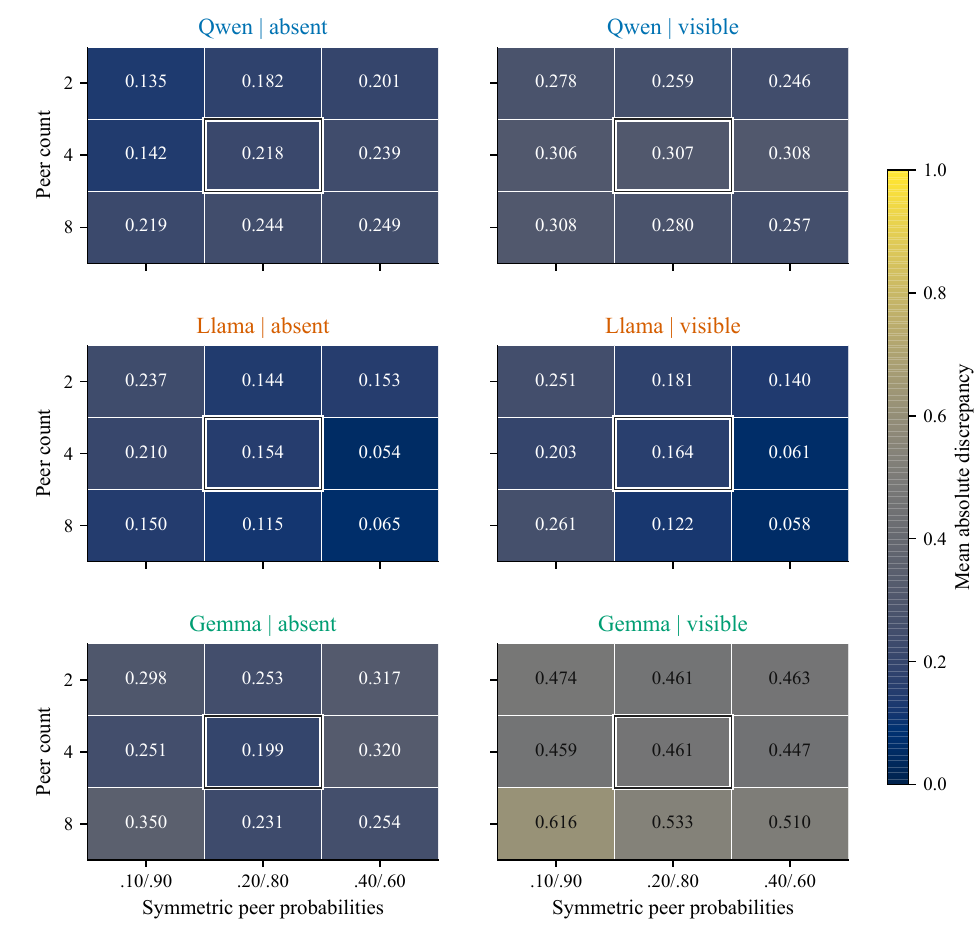}
\caption{Mean absolute complement discrepancy by peer count and symmetric report pair. Each cell averages six worlds, two priors and three opposition levels; the outlined cell is the four-peer, 0.2/0.8 setting used in the main tests.}\label{fig:appendix-M5_peer_grid}\end{figure*}

\paragraph{Peer count and report strength.} Figure~\ref{fig:appendix-M5_peer_grid} gives the full grid. Each cell averages six worlds, two priors and three opposition levels (36 complementary pairs). All 54 cells are positive. Llama's discrepancy generally falls as reports move toward $0.4/0.6$, whereas Gemma's is highest with eight peers at $0.1/0.9$.

\paragraph{Repair by orientation.} Table~\ref{tab:repair} gives the losses behind Figure~\ref{fig:repair}. Symmetrization gives both orientations the same loss because it keeps one population. The reminder brings Qwen's original-orientation loss to zero but leaves its complementary loss at $0.229$, and for Gemma it raises the complementary loss.

\begin{table}[h]
\centering
\small
\setlength{\tabcolsep}{3.5pt}
\begin{tabular}{llrrrr}
\toprule
Model & Arm & $L_0(5)$ & $L_1(5)$ & Final & Rounds \\
\midrule
Qwen  & Baseline        & 0.083 & 0.250 & 0.167 & 0.168 \\
      & Reminder        & 0.000 & 0.229 & 0.115 & 0.138 \\
      & Greedy average  & 0.083 & 0.250 & 0.167 & 0.168 \\
      & Symmetrization  & 0.147 & 0.147 & 0.147 & 0.098 \\
\midrule
Llama & Baseline        & 0.071 & 0.345 & 0.208 & 0.196 \\
      & Reminder        & 0.074 & 0.209 & 0.142 & 0.191 \\
      & Greedy average  & 0.071 & 0.345 & 0.208 & 0.196 \\
      & Symmetrization  & 0.153 & 0.153 & 0.153 & 0.169 \\
\midrule
Gemma & Baseline        & 0.261 & 0.291 & 0.276 & 0.259 \\
      & Reminder        & 0.247 & 0.334 & 0.290 & 0.268 \\
      & Greedy average  & 0.261 & 0.291 & 0.276 & 0.259 \\
      & Symmetrization  & 0.116 & 0.116 & 0.116 & 0.144 \\
\bottomrule
\end{tabular}
\caption{Visible-evidence Brier loss after five rounds of feedback, averaged over six worlds. $L_0(5)$ and $L_1(5)$ are final losses in each orientation, Final is their mean, and Rounds averages that mean over rounds 1--5. Lower is better.}
\label{tab:repair}
\end{table}

\begin{figure*}[t]\centering
\includegraphics[width=\textwidth]{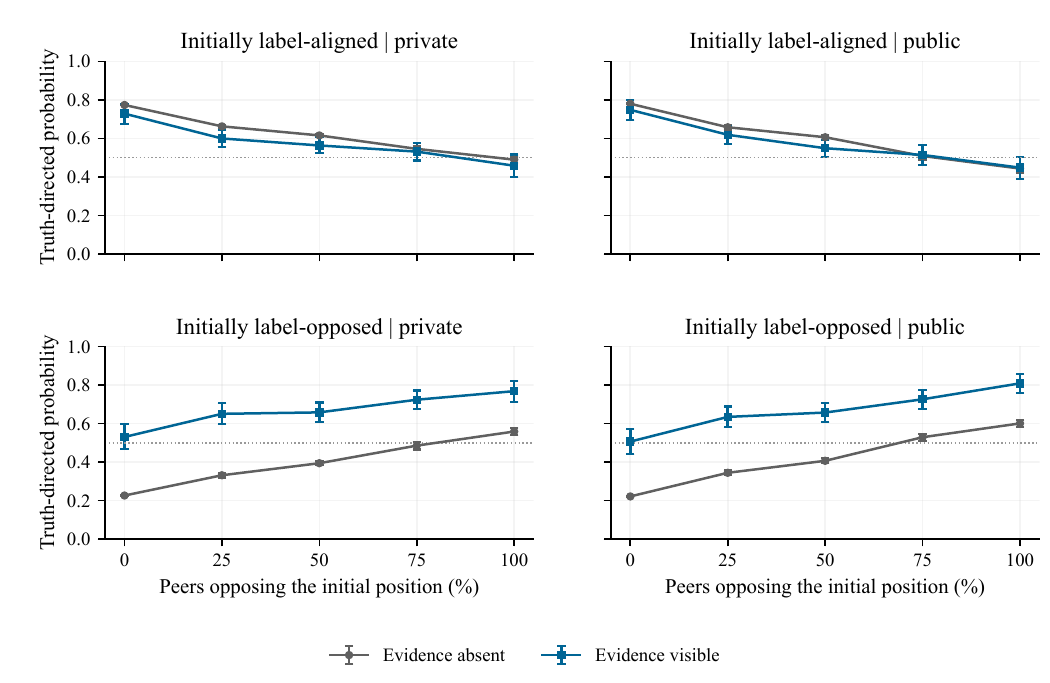}
\caption{Truth-directed probability on the 48 real claims as the share of peers opposing the initial position increases, averaged over claims and both wordings. Rows: initial position aligned with or opposed to the dataset label. Columns: audience. Error bars are 95\% claim-bootstrap intervals; the dotted line marks probability 0.5.}\label{fig:original-dose}\end{figure*}

\paragraph{Real claims by label.} Table~\ref{tab:real-labels} splits the real-claim results by AVeriTeC label. Refuted claims have a small signed discrepancy but a large absolute one, because pairs with opposite signs cancel. Figure~\ref{fig:original-dose} shows how truth-directed probability changes as more peers oppose the initial position. Visible evidence mainly helps agents that start on the wrong side; it does little to protect agents that start on the right side.

\begin{table}[h]
\centering
\small
\setlength{\tabcolsep}{4pt}
\begin{tabular}{llrr}
\toprule
Label & Quantity & Estimate & 95\% interval \\
\midrule
Supported & Signed $\delta$   & 0.3488 & $[0.3016, 0.3948]$ \\
          & Absolute $|\delta|$ & 0.3647 & $[0.3276, 0.4042]$ \\
          & Protection $P$    & 0.0100 & $[-0.0822, 0.1039]$ \\
\midrule
Refuted   & Signed $\delta$   & 0.0246 & $[-0.0673, 0.1163]$ \\
          & Absolute $|\delta|$ & 0.2872 & $[0.2307, 0.3469]$ \\
          & Protection $P$    & 0.0400 & $[-0.0572, 0.1323]$ \\
\bottomrule
\end{tabular}
\caption{Real-claim results by dataset label (24 claims each), with pointwise 95\% intervals from a claim bootstrap within label.}
\label{tab:real-labels}
\end{table}

\section{Appendix D: Implementation Details}
\label{app:details}

All models run in llama.cpp with a 4,096-token context and at most 192 generated tokens. Greedy decoding uses temperature 0, top-$p$ 1, top-$k$ 0, min-$p$ 0 and repetition penalty 1; repeated sampling uses temperature 0.2 with a separate seed per draw. Llama's chat template inserts the current date, which is fixed to the same value in every compared Llama request. Gemma receives the system instruction and user message as one user turn and is constrained with a JSON schema instead of a grammar. A response is accepted only if it contains exactly one finite number in $[0,1]$ and no other field.

In the network study, initial probabilities are drawn uniformly from $[0.2, 0.8]$. Degree targeting breaks ties by a seeded shuffle; random targeting samples three agents without replacement. Each agent's neighbors are shuffled at every update, and response seeds depend on world seed, scenario, round and agent, matched across policies and schedules. Bootstrap intervals resample scenario pairs crossed with world seeds in the network and preserved-state studies, and claims within label in the real-claim study. Peer orders are averaged before any absolute value, loss or flip indicator is computed.

In the feedback studies, each agent's four neighbors on the radius-two ring are sorted by agent index, every round reads the complete previous state, and reports are passed on as recorded, without rounding or smoothing.

\end{document}